\documentclass[a4paper,fleqn]{cas-sc}
\usepackage{amsthm}
\usepackage{booktabs}
\usepackage{array}
\usepackage{hyperref}
\usepackage{overpic}
\usepackage{xcolor}
\usepackage{pifont}

\usepackage{amssymb}
\usepackage{graphicx}
\usepackage{epstopdf}
\usepackage[numbers]{natbib}
\usepackage{bm}
\usepackage{float}
\usepackage{algorithm}
 \usepackage{dsfont}
\usepackage{multirow}
\usepackage{longtable}
\usepackage{subfigure}
\usepackage{supertabular}
\usepackage{algpseudocode}
\usepackage{amsmath}
\usepackage[justification=centering]{caption}

\usepackage{algorithm,float}

\def\tsc#1{\csdef{#1}{\textsc{\lowercase{#1}}\xspace}}
\tsc{WGM}
\tsc{QE}

\begin{document}
\let\WriteBookmarks\relax
\def\floatpagepagefraction{1}
\def\textpagefraction{.001}

\shorttitle{}    

\shortauthors{M. Jiang et al.}  

\title [mode = title]{Multi-View Fair Clustering Guided by Cross-View Sensitive Information Discrepancy}  



%

\author[1]{Mudi Jiang}[orcid=0000-0001-9474-8375]


\ead{32017044@mail.dlut.edu.cn}

\author[1]{Jiahui Zhou}



\author[1]{Xinying Liu}



\author[1]{Zengyou He}


\author[1]{Zhikui Chen}
\cormark[1]

\ead{zkchen@dlut.edu.cn}

\affiliation[1]{organization={School of Software, Dalian University of Technology}, 
            city={Dalian},
            state={Liaoning},
            country={China}}


\cortext[1]{Corresponding author}



\begin{abstract}
Multi-view clustering (MVC) aims to uncover latent cluster structures by exploiting complementary information from multiple views.
Despite substantial progress in clustering performance, fairness remains an important concern when MVC is applied to socially sensitive scenarios.
Recent fair multi-view clustering methods have introduced fairness constraints into representation learning or clustering assignments. However, these methods generally treat different views under a largely uniform fairness mechanism, without explicitly distinguishing their varying levels of sensitive dependence during cross-view learning. In practice, different views may encode substantially different levels of sensitive information. Ignoring such cross-view discrepancy can allow highly sensitive-dependent views to influence less sensitive-dependent ones during cross-view learning, potentially degrading both clustering performance and fairness.
To address this issue, we propose a novel multi-view fair clustering framework guided by cross-view sensitive information discrepancy.
Specifically, we estimate the sensitive dependence of each view and develop a bias-ranked asymmetric alignment mechanism that encourages views with higher sensitive dependence to learn from those with lower sensitive dependence, while cross-view discrepancies are further exploited to adaptively regulate the alignment process.
Moreover, fairness regularization is imposed on the consensus soft assignments to further promote group fairness.
Extensive experiments on benchmark datasets demonstrate that the proposed method achieves a favorable balance between clustering quality and group fairness.
Furthermore, robustness analysis under heterogeneous sensitive dependence demonstrates that the proposed method is more robust to cross-view sensitive-information heterogeneity than existing multi-view fair clustering approaches, maintaining more stable fairness while preserving competitive clustering performance.
\end{abstract}




\maketitle

\section{Introduction}\label{1}
Multi-view data~\cite{liu2025two} describe the same set of instances from multiple perspectives, providing complementary information that cannot be fully captured by a single view.
Such data are particularly valuable for clustering analysis~\cite{oyewole2023data}, where latent structures must be uncovered without label supervision.
Multi-view clustering~\cite{fang2023comprehensive} therefore seeks to integrate complementary information across views to obtain more informative and reliable clustering structures. Existing studies have investigated a wide range of techniques, including subspace learning, matrix factorization, graph-based modeling, and deep representation learning, and have achieved substantial advances in clustering performance. Nevertheless, conventional multi-view clustering methods are primarily designed to optimize clustering effectiveness, with limited consideration of whether the resulting partitions treat different demographic groups equitably. This limitation is particularly critical in socially sensitive applications such as financial services, healthcare, and personnel management, where clustering outcomes may influence downstream decisions and potentially reinforce disparities associated with sensitive attributes.

Fairness has consequently received increasing attention in clustering research~\cite{chhabra2021overview}. Depending on the level at which equitable treatment is defined, existing fairness notions are generally studied from individual and group perspectives. Individual fairness~\cite{jung_et_al:LIPIcs.FORC.2020.5} emphasizes local consistency, requiring instances with similar characteristics to receive comparable clustering outcomes. In contrast, group fairness~\cite{NIPS2017_978fce5b} focuses on the distribution of clustering outcomes across populations defined by sensitive attributes, with the aim of preventing specific demographic groups from being systematically favored or disadvantaged. This notion is closely related to the principle of disparate impact~\cite{rutherglen1987disparate}, which concerns disparities in outcomes across protected groups even in the absence of explicitly discriminatory rules. Since many practical clustering applications are primarily concerned with population-level disparities, this work focuses on group fairness in the multi-view setting.

In recent years, fairness has received increasing attention in multi-view clustering~\cite{zheng2023fairness,zhao2024dfmvc,li2024one,Xu_2025_CVPR,zhao2026fairness,jiang2026adversarial}. Existing methods mainly extend conventional multi-view clustering frameworks by introducing fairness regularization into representation learning or cluster assignments. However, they generally do not explicitly distinguish the varying levels of sensitive dependence across views, although different views may encode considerably different amounts of sensitive information. When such heterogeneous views are fused or encouraged to interact without accounting for these differences, information associated with sensitive attributes from views with higher sensitive dependence may influence those with lower sensitive dependence, potentially weakening low-dependence representations and compromising both clustering quality and fairness.

To address the above issue, we propose a novel multi-view fair  clustering framework that explicitly models cross-view discrepancies in sensitive information and leverages them to guide the training process. Specifically, the framework first employs view-specific autoencoders to learn latent representations for individual views, based on which the dependence between each view and the sensitive attribute is quantified. These estimates are then used to guide an asymmetric alignment mechanism, in which views with higher sensitive dependence are encouraged to learn from those with lower sensitive dependence. Meanwhile, pairwise and global cross-view discrepancies are further exploited to determine the relative importance and adaptive strength of the alignment. In this manner, cross-view interaction is explicitly regulated according to the heterogeneous levels of sensitive information encoded by different views, thereby mitigating the propagation of sensitive information while preserving useful complementary structures. Finally, fairness regularization is imposed on the consensus soft assignments aggregated across views to further promote group fairness at the clustering-assignment level. By jointly optimizing representation learning, discrepancy-guided asymmetric alignment, and assignment-level fairness, the proposed framework seeks to achieve a favorable trade-off between clustering quality and fairness.

To evaluate the proposed framework, we conduct extensive experiments on five benchmark multi-view datasets with fairness constraints. The results show that our method achieves a favorable balance between clustering quality and group fairness, with competitive performance compared with state-of-the-art multi-view clustering and fairness-aware clustering methods. Moreover, the robustness analysis under heterogeneous view bias demonstrates that the proposed framework can better cope with cross-view sensitive-information heterogeneity, maintaining more stable fairness while preserving competitive clustering performance under increasing sensitive dependence in the target view.

The main contributions of this paper are summarized as follows:
\begin{itemize}
    \item We first investigate the cross-view sensitive information discrepancy problem in multi-view fair clustering, explicitly considering the heterogeneous dependence of different views on sensitive attributes.

    \item We develop a joint optimization framework that integrates discrepancy-guided asymmetric view alignment, multi-view clustering consistency learning, and fairness regularization on consensus assignments, jointly promoting discriminative representations, consistent clustering structures, and group-fair assignments.

    \item Extensive experiments on five benchmark datasets with fairness constraints demonstrate that the proposed method achieves a favorable balance between clustering quality and group fairness. Moreover, the robustness analysis under heterogeneous view bias further shows that our method is better able to handle cross-view sensitive-information heterogeneity than existing fair multi-view clustering methods, maintaining more stable fairness while preserving competitive clustering performance under increasing sensitive dependence in the target view.
\end{itemize}

The rest of this paper is organized as follows. Section~\ref{2} summarizes the related studies. Section~\ref{3} presents the proposed method. Section~\ref{4} provides the experimental results. Finally, Section~\ref{5} concludes this work.

\section{Related work}\label{2}
\subsection{Fair Clustering}
\subsubsection{Individual fairness}
Individual fairness considers fairness from the perspective of local consistency among data instances. Its underlying principle is that samples exhibiting similar characteristics in the original feature space should receive comparable clustering outcomes, regardless of their sensitive-group memberships. Rather than assessing fairness solely at the population level, individual fairness imposes constraints on the relative treatment of neighboring or structurally similar instances. Jung et al.~\cite{jung_et_al:LIPIcs.FORC.2020.5} introduced an early formulation based on distance-aware fairness constraints that relate clustering decisions to the local structure of the data. Building on this foundation, subsequent studies developed bicriteria approximation algorithms that jointly control clustering quality and fairness violations~\cite{pmlr-v119-mahabadi20a}. Later work further improved the approximation guarantees for different clustering objectives, including $k$-means and $k$-median~\cite{10.5555/3540261.3541283,pmlr-v151-vakilian22a}. Meanwhile, several studies have focused on reducing the computational complexity of individually fair clustering, thereby improving its applicability to large-scale datasets~\cite{pmlr-v151-chhaya22a,pmlr-v238-bateni24a}.

\subsubsection{Group fairness}

Group fairness, in contrast, focuses on disparities among populations defined by sensitive attributes, such as race or gender. Its primary objective is to prevent the resulting clusters from exhibiting systematically imbalanced or discriminatory distributions across sensitive groups. Existing approaches generally incorporate fairness at different stages of the clustering pipeline.

Pre-processing methods modify or augment the input data prior to clustering so that the transformed data better satisfy fairness requirements. Representative strategies include fairlet-based decomposition, which partitions the dataset into small balanced subsets~\cite{NIPS2017_978fce5b,NEURIPS2020_f10f2da9}, and fair coreset construction, which compresses the original dataset while preserving both clustering structure and fairness properties~\cite{schmidt2020fair,NEURIPS2019_810dfbbe}. Other studies introduce carefully designed antidote samples to steer the subsequent clustering process toward fairer solutions~\cite{pmlr-v171-chhabra22a}.

In-processing approaches incorporate fairness directly into the clustering objective or optimization procedure. This category includes fairness-constrained extensions of classical clustering algorithms, such as spectral clustering and $k$-median clustering~\cite{kleindessner2019guarantees,10.1007/978-3-030-86520-7_47}, as well as deep clustering frameworks that jointly optimize clustering quality and fairness through adversarial learning or multi-objective optimization~\cite{zhang2021deep,li2020deep}. In contrast, post-processing methods leave the original clustering procedure unchanged and instead refine the obtained solution, for example by reassigning samples or adjusting cluster centers to improve fairness~\cite{kleindessner2019fair,jones2020fair}.

Overall, existing fair clustering research has developed a broad spectrum of mechanisms for promoting fairness at either the individual or group level. However, most existing methods are primarily designed for single-view data, and their applicability to multi-view clustering remains relatively limited.

\subsection{Multi-View Clustering}
\subsubsection{Traditional multi-view clustering}
Multi-view clustering aims to uncover a common clustering structure by jointly exploiting the complementary and shared information embedded in multiple views. Existing studies have addressed this problem from various perspectives. Graph-based methods typically construct view-specific similarity graphs to characterize neighborhood or structural relationships within each view and subsequently integrate them into a unified representation for clustering~\cite{li2021consensus,liang2019consistency}. Another representative line of research is subspace learning, which projects heterogeneous views into a shared low-dimensional space to enhance common information while suppressing view-specific noise~\cite{li2019flexible,yang2019split}. Matrix factorization-based methods pursue a similar objective by decomposing multi-view observations into compact latent factors, from which the underlying clustering structure can be inferred~\cite{wang2018multiview,yang2020uniform}.
To capture more complex relationships among samples, kernel-based methods employ nonlinear mappings and integrate multiple kernel representations to model cross-view dependencies~\cite{liu2023contrastive,yuan2022robust}. More recently, deep multi-view clustering has attracted increasing attention. By leveraging neural networks to learn nonlinear high-level representations, these methods enable representation learning and clustering to be optimized in a unified manner, offering greater flexibility in modeling heterogeneous multi-view data~\cite{10.5555/3367243.3367449,gao2020cross}. 

Despite their effectiveness in exploiting cross-view complementarity, most conventional multi-view clustering methods primarily focus on improving clustering consistency and representation quality, while fairness is rarely considered as an explicit learning objective. Consequently, the learned representations and clustering assignments may still reflect information associated with sensitive attributes, potentially leading to unequal clustering outcomes across different demographic groups. This limitation has motivated increasing research interest in incorporating fairness considerations into the multi-view clustering process.
\subsubsection{Fairness-related multi-view clustering}
Recent studies have extended multi-view clustering by explicitly incorporating fairness into representation learning, cluster assignment, or multi-view fusion. Fair-MVC~\cite{zheng2023fairness} introduces group-level fairness constraints by encouraging balanced proportions of protected groups across clusters. DFMVC~\cite{zhao2024dfmvc} further integrates contrastive representation learning with fairness-aware assignment regularization, guiding subgroup distributions toward a predefined target while preserving cross-view consistency. Xu et al.~\cite{Xu_2025_CVPR} enhance fair representation learning by incorporating Kolmogorov--Arnold networks into the multi-view clustering framework, with the aim of capturing more complex feature dependencies while improving both robustness and fairness.

Other studies promote fairness through clustering structures or optimization mechanisms. FMSC~\cite{li2024one} develops a spectral clustering formulation that encourages group fairness by regulating the connectivity patterns of protected-group subgraphs within clusters. AFMVC~\cite{jiang2026adversarial} adopts an adversarial learning strategy to suppress sensitive information in view-specific latent representations, employing a discriminator with gradient reversal to reduce the predictability of sensitive attributes while preserving clustering quality. More recently, FLFMVC \cite{zhao2026fairness} introduces a fairness-aware late-fusion framework that jointly considers robustness, demographic balance, and fair cluster assignment through fairness-aware graph filtering, group-normalized regularization, and fairness-guided discretization.

Despite these advances, existing multi-view fair clustering methods mainly introduce fairness constraints into conventional multi-view clustering frameworks, while paying limited attention to the heterogeneity of sensitive information across views. In practice, different views may exhibit markedly different degrees of dependence on sensitive attributes, and conventional fusion or consistency learning may therefore facilitate the propagation of sensitive information across views, potentially compromising both clustering quality and fairness. Motivated by this limitation, we explicitly model cross-view discrepancies in sensitive information to guide asymmetric view alignment, while further imposing fairness regularization on the consensus assignments.
\section{Method}
\label{3}
\begin{figure}[pos=H]
\centering
\includegraphics[width=1\textwidth]{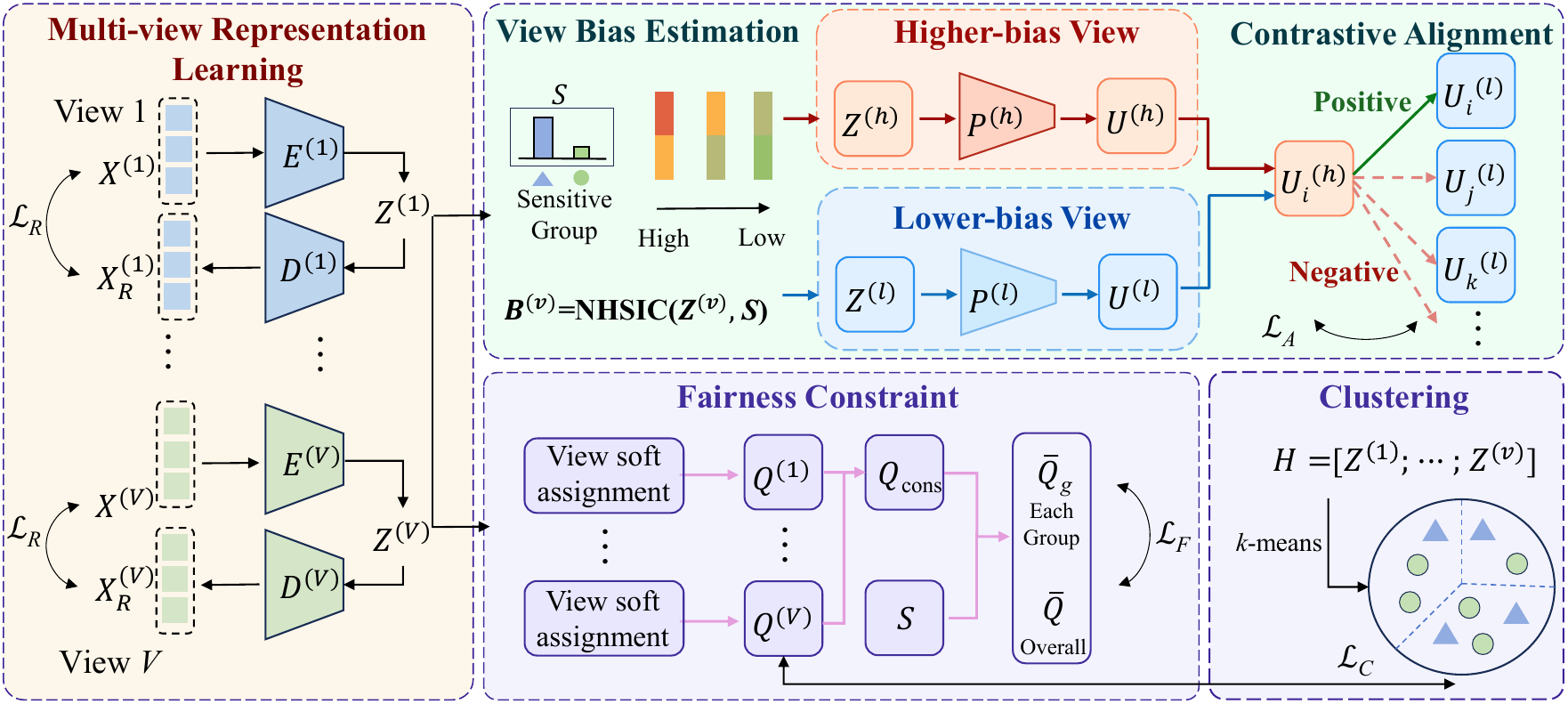} 
\caption{Overview of the proposed multi-view fair clustering framework.}
\label{fig1}
\end{figure}
In this section, we introduce the overall architecture of the proposed multi-view fair  clustering framework. As illustrated in Fig.~\ref{fig1}, the framework comprises three main components: Multi-view Representation Learning, Bias-Ranked Asymmetric View Alignment, and Fairness Regularization on Consensus Assignments. The first component learns informative view-specific representations to capture complementary clustering structures across multiple views. The second explicitly accounts for heterogeneous sensitive dependence across views and performs asymmetric contrastive alignment, whereby views with higher sensitive dependence are guided by those with lower sensitive dependence. The third further enhances fairness at the assignment level by imposing a demographic parity constraint on the consensus soft assignments aggregated across views. Finally, the learned view-specific representations are aggregated to derive the final clustering assignments.

\subsection{Notations and Problem Definition}
Let $\mathcal{X}=\{X^{(v)}\}_{v=1}^{V}$ denote a multi-view dataset comprising $N$ instances observed from $V$ distinct views, where $X^{(v)}\in\mathbb{R}^{N\times d_v}$ is the feature matrix of the $v$-th view and $d_v$ denotes its feature dimensionality. Correspondingly, $X_i^{(v)}$ represents the feature vector of the $i$-th instance in view $v$. Each instance is further associated with a sensitive attribute $S_i\in\mathcal{S}$, where $\mathcal{S}$ denotes the set of sensitive groups. Given a predefined number of clusters $K$, the objective of multi-view fair clustering is to exploit the complementary information across multiple views to uncover meaningful cluster structures while mitigating disparities among different sensitive groups.
\subsection{Multi-View Representation Learning and Clustering}

To capture complementary information across multiple views while preserving their view-specific characteristics, we employ an independent autoencoder for each view. Specifically, the encoder $E^{(v)}$ maps $X^{(v)}$ into a latent representation:
\begin{equation}
\label{eq:encoder}
Z^{(v)}
=
E^{(v)}
(
X^{(v)};\theta_E^{(v)}
),
\end{equation}
where $Z^{(v)} \in \mathbb{R}^{N \times d_z}$ denotes the latent representation of view $v$, and $\theta_E^{(v)}$ denotes the parameters of the corresponding encoder.

The resulting latent representation is then fed into a view-specific decoder $D^{(v)}$ to reconstruct the original input:
\begin{equation}
\label{eq:decoder}
X_R^{(v)}
=
D^{(v)}
(
Z^{(v)};\theta_D^{(v)}
),
\end{equation}
where $X_R^{(v)}$ denotes the reconstructed feature matrix and $\theta_D^{(v)}$ denotes the parameters of the corresponding decoder. To retain the intrinsic information contained in each view, we minimize the reconstruction discrepancy between the original and reconstructed features:
\begin{equation}
\label{eq:reconstruction}
\mathcal{L}_R
=
\sum_{v=1}^{V}
\left\|
X^{(v)}-X_R^{(v)}
\right\|_F^2.
\end{equation}
Before joint optimization, the view-specific autoencoders are initialized through reconstruction pretraining, which provides a stable representation space for subsequent clustering and fairness learning.

Based on the learned latent representations, we further construct a shared clustering structure across views. Specifically, the latent representations from all views are concatenated to form a unified multi-view representation:
\begin{equation}
\label{eq:fused_representation}
H
=
[
Z^{(1)};
Z^{(2)};
\cdots;
Z^{(V)}
].
\end{equation}
$k$-means is then applied to  obtain global pseudo cluster labels, which provide a common clustering target for all views. Let
$Y \in \{0,1\}^{N \times K}$
denote the corresponding one-hot pseudo-label matrix, where $Y_{ik}=1$ indicates that instance $i$ is assigned to cluster $k$. During training, the pseudo labels are periodically updated according to the current multi-view representation, allowing the shared clustering structure to progressively adapt to the learned feature space.

For each view $v$, we introduce $K$ learnable cluster centroids
$\{\mu_k^{(v)}\}_{k=1}^{K}$ and construct a view-specific soft assignment matrix
$Q^{(v)} \in \mathbb{R}^{N \times K}$ using a Student's $t$-distribution kernel:
\begin{equation}
\label{eq:student_t}
Q_{ik}^{(v)}
=
\frac{
(
1+
\|
Z_i^{(v)}-\mu_k^{(v)}
\|_2^2/\alpha
)^{-(\alpha+1)/2}
}{
\displaystyle
\sum_{k'=1}^{K}
(
1+
\|
Z_i^{(v)}-\mu_{k'}^{(v)}
\|_2^2/\alpha
)^{-(\alpha+1)/2}
}.
\end{equation}
Here, $Q_{ik}^{(v)}$ represents the soft probability of assigning instance $i$ to cluster $k$ in view $v$, and $\alpha$ denotes the degree-of-freedom parameter of the Student's $t$-distribution, which is set to $1$ in our experiments. The view-specific cluster centroids are initialized by applying $k$-means to the corresponding latent representations and are subsequently optimized jointly with the network parameters.

To encourage each view to preserve a clustering structure consistent with the shared multi-view partition, we minimize the KL divergence between the global pseudo target $Y$ and each view-specific soft assignment $Q^{(v)}$:
\begin{equation}
\label{eq:clustering_loss}
\mathcal{L}_C
=
\sum_{v=1}^{V}
\sum_{i=1}^{N}
\sum_{k=1}^{K}
Y_{ik}
\log
(
Y_{ik}/Q_{ik}^{(v)}
).
\end{equation}

By minimizing $\mathcal{L}_C$, the fused representation $H$ captures the shared clustering structure by integrating information from all views, while $Q^{(v)}$ characterizes the soft clustering behavior of each individual view. The learned latent representations $Z^{(v)}$ are subsequently used to estimate cross-view sensitive information discrepancy and guide asymmetric view alignment, while the soft assignments $Q^{(v)}$ are aggregated to construct the consensus assignment for fairness regularization.

\subsection{Asymmetric View Alignment Guided by Sensitive Information Discrepancy}

Different views may exhibit varying degrees of dependence on the sensitive attribute.
Therefore, directly enforcing cross-view consistency without accounting for such differences may allow sensitive information from views with higher sensitive dependence to influence those with lower sensitive dependence.
To address this issue, we estimate the sensitive dependence of each view and leverage the resulting cross-view discrepancy to guide asymmetric representation alignment.

Given the latent representation $Z^{(v)}$ and the sensitive attribute $S$, we quantify the sensitive information contained in view $v$ using the normalized Hilbert--Schmidt Independence Criterion (NHSIC):
\begin{equation}
\label{eq:view_bias}
B^{(v)}
=
\operatorname{NHSIC}
(
Z^{(v)}, S
),
\end{equation}
where a larger $B^{(v)}$ indicates stronger statistical dependence between the representation of view $v$ and the sensitive attribute.
The resulting sensitive-dependence score is used to characterize the relative amount of sensitive information across views.

To perform cross-view alignment without directly constraining the clustering representation space, each latent representation is further mapped into a view-specific projection space:
\begin{equation}
\label{eq:projection}
U^{(v)}
=
P^{(v)}
(
Z^{(v)}
),
\end{equation}
where $P^{(v)}$ denotes the projection head associated with view $v$.
For any pair of views $(h,l)$ satisfying
\begin{equation}
B^{(h)} > B^{(l)},
\end{equation}
view $h$ is regarded as having relatively higher sensitive dependence, while view $l$ exhibits lower sensitive dependence.
Accordingly, we align the former toward the latter rather than treating the two views equally during cross-view alignment.
To explicitly preserve this directional relationship, the representation of the lower-dependence view is treated as a fixed reference:
\begin{equation}
\label{eq:stop_gradient}
\bar{U}^{(l)}
=
\operatorname{sg}
(
U^{(l)}
),
\end{equation}
where $\operatorname{sg}(\cdot)$ denotes the stop-gradient operation.
The directed contrastive alignment between views $h$ and $l$ is then defined as
\begin{equation}
\label{eq:pair_alignment}
\mathcal{L}_{h\rightarrow l}
=
-\frac{1}{N}
\sum_{i=1}^{N}
\log
\frac{
e^{
(
\operatorname{sim}
(
U_i^{(h)}, \bar{U}_i^{(l)}
)/\tau
)
}}{
\displaystyle
\sum_{j=1}^{N}
e^{
(
\operatorname{sim}
(
U_i^{(h)}, \bar{U}_j^{(l)}
)/\tau
)
}},
\end{equation}
where $\operatorname{sim}(\cdot,\cdot)$ denotes cosine similarity and $\tau$ is the temperature parameter.
Representations of the same instance across the two views constitute positive pairs, whereas those of different instances are treated as negative pairs.
By detaching the lower-dependence branch, only the higher-dependence branch is updated toward the reference representation.

Moreover, different view pairs may exhibit different levels of sensitive information discrepancy.
For each directed pair $(h,l)$, we therefore define a discrepancy-aware weight as
\begin{equation}
\label{eq:pair_weight}
w_{hl}
=
\frac{
B^{(h)}-B^{(l)}
}{
\displaystyle
\sum_{(p,q)\in\mathcal{A}}
\left(
B^{(p)}-B^{(q)}
\right)
},
\end{equation}
where
\begin{equation}
\label{eq:alignment_pairs}
\mathcal{A}
=
\left\{
(h,l)\mid B^{(h)}>B^{(l)}
\right\}
\end{equation}
denotes the set of all directed high-to-low view pairs.
Accordingly, view pairs with larger sensitive information discrepancies receive greater emphasis during alignment.
The asymmetric alignment objective before adaptive calibration is formulated as
\begin{equation}
\label{eq:alignment_loss}
\mathcal{L}_A^{\mathrm{raw}}
=
\sum_{(h,l)\in\mathcal{A}}
w_{hl}
\mathcal{L}_{h\rightarrow l}.
\end{equation}

To further characterize the overall heterogeneity of sensitive information across views, we define
\begin{equation}
\label{eq:global_discrepancy}
g
=
\min
\left(
\frac{
B_{\max}-B_{\min}
}{
\bar{B}+\epsilon
},
1
\right),
\end{equation}
where
\begin{equation}
\bar{B}
=
\frac{1}{V}
\sum_{v=1}^{V}
B^{(v)},
\end{equation}
with $B_{\max}=\max_v B^{(v)}$ and $B_{\min}=\min_v B^{(v)}$.
Here, $\epsilon$ is a small positive constant for numerical stability.
A larger $g$ indicates greater heterogeneity in sensitive information across views.

During optimization, $g$ is further used to adaptively calibrate the effective contribution of the alignment objective relative to the clustering objective.
A larger cross-view sensitive-information discrepancy leads to stronger alignment, whereas the alignment effect is reduced when different views exhibit similar levels of sensitive dependence.
The resulting calibrated alignment objective is denoted as $\mathcal{L}_A$.
In this way, cross-view sensitive information discrepancy determines the alignment direction, the relative importance of different view pairs, and the effective strength of representation alignment.

\subsection{Fairness Regularization}

Although the preceding asymmetric alignment mitigates sensitive information at the representation level, such representation-level regulation does not necessarily guarantee fairness in the final clustering assignments.
In particular, residual sensitive dependence may still be reflected in the aggregated cluster distribution across different groups.
To explicitly promote group fairness, we introduce an additional fairness regularization term based on the aggregated soft assignments across multiple views.

Given the view-specific soft assignment matrices $\{Q^{(v)}\}_{v=1}^{V}$, we construct the consensus assignment as
\begin{equation}
\label{eq:consensus_assignment}
Q_{\mathrm{cons}}
=
\frac{1}{V}
\sum_{v=1}^{V}
Q^{(v)}.
\end{equation}
Here, $Q_{\mathrm{cons}} \in \mathbb{R}^{N\times K}$ represents the aggregated clustering assignment shared across all views.

Let $\mathcal{S}$ denote the set of sensitive groups and
\begin{equation}
\mathcal{I}_g
=
\left\{
i \mid S_i=g
\right\}
\end{equation}
denote the set of instances belonging to group $g$, with $N_g=|\mathcal{I}_g|$.
For each sensitive group $g$ and cluster $k$, the average assignment probability is defined as
\begin{equation}
\label{eq:group_assignment}
\bar{Q}_{gk}
=
\frac{1}{N_g}
\sum_{i\in\mathcal{I}_g}
Q_{\mathrm{cons},ik},
\end{equation}
while the overall average assignment probability for cluster $k$ is
\begin{equation}
\label{eq:overall_assignment}
\bar{Q}_{k}
=
\frac{1}{N}
\sum_{i=1}^{N}
Q_{\mathrm{cons},ik}.
\end{equation}
Based on these soft assignments, we encourage each sensitive group to exhibit a cluster distribution close to the overall population distribution.
Accordingly, the fairness regularization is formulated as
\begin{equation}
\label{eq:fairness_loss}
\mathcal{L}_F
=
\frac{1}{|\mathcal{S}|K}
\sum_{g\in\mathcal{S}}
\sum_{k=1}^{K}
\left(
\bar{Q}_{gk}
-
\bar{Q}_{k}
\right)^2.
\end{equation}

By minimizing $\mathcal{L}_F$, the expected assignment probability to each cluster is encouraged to be consistent across different sensitive groups, providing a differentiable relaxation of demographic parity at the clustering-assignment level.
Importantly, this fairness constraint is imposed on the consensus assignments rather than directly on individual latent representations, allowing the view-specific representations to retain information useful for clustering while jointly promoting fairness in the final partition.

\subsection{Overall Objective Function}

The proposed framework jointly considers representation preservation, clustering consistency, discrepancy-guided cross-view alignment, and assignment-level fairness.
Accordingly, the overall objective is formulated as
\begin{equation}
\label{eq:overall_objective}
\mathcal{L}
=
\mathcal{L}_R
+
\lambda_C \mathcal{L}_C
+
\mathcal{L}_A
+
\lambda_F \mathcal{L}_F,
\end{equation}
where $\mathcal{L}_R$ preserves the intrinsic information of each view through reconstruction, $\mathcal{L}_C$ encourages view-specific assignments to follow the shared clustering structure, $\mathcal{L}_A$ regulates cross-view representation alignment according to sensitive information discrepancy, and $\mathcal{L}_F$ promotes demographic parity on the consensus assignments. 
The hyperparameters $\lambda_C$ and $\lambda_F$ control the contributions of the clustering and fairness objectives, respectively, while the strength of $\mathcal{L}_A$ is adaptively determined by the cross-view sensitive information discrepancy as described above. The overall training procedure is shown in Algorithm \ref{alg:framework}.
\begin{algorithm}[ht]
\caption{Training Procedure of the Proposed Framework}
\label{alg:framework}
\textbf{Input}: Multi-view data $\mathcal{X}=\{X^{(v)}\}_{v=1}^{V}$;
sensitive attribute $S$;
number of clusters $K$;
trade-off parameters $\lambda_C$ and $\lambda_F$;
temperature parameter $\tau$;
number of training epochs $T$.\\
\textbf{Output}: Final clustering assignments.
\begin{algorithmic}[1]

\State Pre-train the view-specific autoencoders by minimizing $\mathcal{L}_R$.

\State Obtain the initial latent representations
$\{Z^{(v)}\}_{v=1}^{V}$ and initialize the view-specific cluster centroids.

\For{$t=1$ to $T$}

    \State Encode each view to obtain $Z^{(v)}$ and construct
    $H$ using Eq. \ref{eq:fused_representation}.

    \State Periodically apply $k$-means to $H$ to update the global pseudo labels $Y$.

    \State Compute the view-specific soft assignments $Q^{(v)}$ and the clustering loss $\mathcal{L}_C$ using Eq. \ref{eq:clustering_loss}.

    \State Estimate the sensitive dependence $B^{(v)}$ of each view.

    \State Construct directed  view pairs according to $B^{(v)}$ and compute the asymmetric alignment loss $\mathcal{L}_A$ using Eqs. \ref{eq:alignment_loss} and \ref{eq:global_discrepancy}.

    \State Aggregate $\{Q^{(v)}\}_{v=1}^{V}$ into $Q_{\mathrm{cons}}$ and compute the fairness loss $\mathcal{L}_F$ using Eq. \ref{eq:fairness_loss}.

    \State Compute the overall objective according to Eq.~\ref{eq:overall_objective} and update the network parameters.

\EndFor

\State Apply $k$-means to $H$ to obtain the final clustering assignments.

\end{algorithmic}
\end{algorithm}

\section{Experiments}
\label{4}

\subsection{Experimental Setup}
\subsubsection{Datasets}
We evaluate the proposed method on five benchmark datasets commonly used in multi-view fair clustering, including Credit, Bank, Law, Mfeat,  and COIL. The detailed statistics of these datasets are provided in Table \ref{dataset}. For the single-view datasets Credit, Bank, and Law, we follow \cite{zheng2023fairness} to construct two complementary views by applying nonlinear transformations, such as Sigmoid and ReLU, to the original feature representations. For the multi-view datasets Mfeat and COIL, we adopt the experimental setting in \cite{pmlr-v54-zafar17a} and generate binary sensitive attributes by independently assigning each sample to one of two sensitive groups according to a Bernoulli distribution with parameter 0.5. To ensure a consistent and computationally feasible comparison across all competing methods, we further randomly sample 5,000 instances from Credit and 10,000 instances from Law, whose original sample sizes are 29,537 and 18,692, respectively. This subsampling strategy is necessary because several baseline methods, including FMSC and MCPL, fail to produce clustering results on datasets of this scale within feasible computational resources.

\begin{table}[pos=H]
\small
	\caption{The summary statistics on datasets used in the performance evaluation.}
	\label{dataset}
	\centering
        
	\begin{tabular}{ccccc}
		\toprule
		Dataset &  \#Samples &  \#Clusters &\#Features & Sensitive Attribute \\
		\midrule
		Credit &  5000   &  5   &  22/22   &   Gender      \\
            Bank &  2907   &  2   &  12/12   &   Marital      \\
            Law  &  10000   &  2       &  10/10   &  Gender   \\
		Mfeat &  2000   &  10   &  216/76/64/6/240/47   &   Synthetic Binary    \\
		COIL &  1440   &  20      &  1021/3304/6750  &  Synthetic Binary  \\
		\bottomrule
	\end{tabular}
\end{table}
\subsubsection{Evaluation metrics}
We evaluate clustering performance using two widely adopted metrics, Clustering Accuracy (ACC) and Normalized Mutual Information (NMI). Both metrics measure the consistency between the obtained clustering assignments and the corresponding ground-truth labels. Larger values of ACC and NMI indicate better clustering performance.

To evaluate group fairness, we consider Balance (BAL) \cite{zheng2023fairness} and Demographic Parity Difference (DPD)\footnote{\url{https://fairlearn.org}} as group-level fairness measures. BAL quantifies the degree to which different sensitive groups are proportionally represented within each cluster, whereas DPD measures the disparity in clustering outcomes across sensitive groups. A higher BAL value indicates a more balanced group distribution, while a lower DPD value corresponds to a smaller demographic disparity and, consequently, better fairness.

\subsubsection{Comparison methods}
To comprehensively evaluate the proposed method, we compare it with representative approaches from three categories: single-view fair clustering, state-of-the-art multi-view clustering, and multi-view fair clustering. Since existing multi-view fair clustering methods remain relatively limited, we additionally include the first two categories to provide broader comparisons in terms of clustering performance, fairness, and their trade-off.

For single-view fair clustering methods, features from different views are concatenated into a unified representation before clustering, allowing these methods to be directly applied in the multi-view setting. BFKM \cite{pan2023balanced} extends the conventional $k$-means objective by incorporating fairness and balance constraints as penalty terms and solves the resulting optimization problem via coordinate descent. VFC \cite{ziko2021variational} adopts a variational formulation in which group disparity is characterized through a KL-divergence-based penalty, thereby enabling a flexible balance between clustering utility and demographic parity. FFC \cite{pan2023fairness} employs a multi-stage optimization strategy consisting of fairness-aware initialization, relaxed constrained optimization, and local refinement. FairDen \cite{krieger2025fairden} incorporates group-level fairness constraints into a density-based spectral clustering framework constructed from density-connectivity distances, enabling the identification of arbitrarily shaped clusters while preserving group fairness.

State-of-the-art multi-view clustering methods are designed to exploit complementary and consistent information across multiple views without explicitly incorporating fairness constraints. These methods are included to evaluate the clustering capability of the proposed approach and to examine the fairness characteristics of conventional multi-view clustering methods. MCPL \cite{cai2024multi} leverages pseudo-label information and latent graph structures to learn discriminative representations, followed by a label fusion mechanism for obtaining the final clustering assignments. CGL \cite{li2021consensus} jointly exploits spectral embedding and low-rank tensor learning to construct a consensus graph that captures shared structural information across views. 3MC \cite{chen2025multi} utilizes deep representations extracted from multiple encoder layers and develops a multi-layer, multi-level contrastive learning framework that jointly performs inter-view feature contrastive learning and inter-layer semantic label contrastive learning.

\begin{table}[pos=H]
\caption{Comparison of various clustering methods in terms of two accuracy metrics and two fairness metrics. For each metric across the datasets, the best score is shown in bold, and the second-best is underlined. A dash ``--'' indicates unavailable results: FairMVC is restricted to two-view datasets (Credit, Bank, Law), while FFC fails to produce results on Mfeat and COIL due to excessive memory usage on high-dimensional data.}
\label{compare}
\renewcommand{\arraystretch}{0.85}
\centering
\setlength{\tabcolsep}{4mm}
\begin{tabular}{c|ccccccc}
\toprule
Method & Metric & Credit & Bank  & Law & Mfeat & COIL & Mean Value  \\
\midrule
& ACC
&0.370
&0.635
&0.550
&0.797
&0.689
&0.608
 \\

BFKM & NMI
&0.188
&0.056
&\textbf{0.081}
&0.754
&0.784
&0.373
 \\

& BAL
&0.348
&0.289
&0.430
&0.432
&\textbf{0.371}
&0.374
 \\

& DPD $\downarrow$
&0.028
&\underline{0.066}
&0.003
&\textbf{0.009}
&\textbf{0.009}
&\underline{0.023}
 \\

\midrule
& ACC
&0.370
&0.633
&0.542
&0.925
&0.750
&0.644
 \\

VFC & NMI
&0.186
&0.056
&0.063
&0.856
&0.821
&0.396
 \\

& BAL
&0.341
&0.289
&0.428
&0.434
&0.347
&0.368
 \\

& DPD $\downarrow$
&0.028
&0.097
&0.011
&0.014
&0.011
&0.032
 \\

\midrule
& ACC
&0.368
&0.633
&0.549
&--
&--
&--
 \\

FFC & NMI
&0.187
&0.056
&\textbf{0.081}
&--
&--
&--
 \\

& BAL
&0.340
&0.286
&0.428
&--
&--
&--
 \\

& DPD $\downarrow$
&0.031
&0.081
&\underline{0.001}
&--
&--
&--
 \\

\midrule
& ACC
&0.420
&0.638
&0.802
&0.554
&\textbf{0.823}
&0.647
 \\

FairDen & NMI
&0.046
&0.062
&0.011
&0.595
&\textbf{0.905}
&0.324
\\

& BAL
&0.343
&0.271
&0.400
&0.240
&0.318
&0.314
 \\

& DPD $\downarrow$
&0.036
&0.097
&0.063
&0.011
&\underline{0.010}
&0.043
 \\

\midrule
& ACC
&0.330
&0.672
&0.839
&0.831
&0.699
&0.674
 \\

MCPL & NMI
&0.104
&0.068
&0.050
&0.814
&0.831
&0.373
 \\

& BAL
&0.356
&0.288
&0.417
&0.404
&0.347
&0.362
\\

& DPD $\downarrow$
&0.021
&0.091
&0.064
&\underline{0.010}
&\underline{0.010}
&0.039
 \\

\midrule
& ACC
&0.368
&0.641
&\textbf{0.888}
&\textbf{0.997}
&0.599
&0.699
 \\

CGL & NMI
&0.123
&0.061
&0.078
&\textbf{0.994}
&0.818
&\underline{0.415}
 \\

& BAL
&0.355
&0.285
&0.362
&\textbf{0.450}
&0.347
&0.360
\\

& DPD $\downarrow$
&0.028
&0.094
&0.064
&\underline{0.010}
&\underline{0.010}
&0.041
 \\

\midrule
& ACC
&0.354
&0.635
&0.563
&0.649
&0.501
&0.540
 \\

3MC & NMI
&0.174
&0.059
&0.048
&0.680
&0.659
&0.324
 \\

& BAL
&0.344
&0.285
&0.420
&0.438
&0.339
&0.365
 \\

& DPD $\downarrow$
&0.032
&0.094
&0.066
&\underline{0.010}
&0.011
&0.043
 \\

\midrule
& ACC
&0.402
&0.623
&0.588
&--
&--
&--
 \\

FairMVC & NMI
&0.117
&0.033
&0.077
&--
&--
&--
 \\

& BAL
&0.341
&0.288
&0.424
&--
&--
&--
 \\

& DPD $\downarrow$
&0.029
&0.092
&0.096
&--
&--
&--
 \\

\midrule
& ACC
&0.387
&0.652
&\underline{0.862}
&0.293
&\underline{0.806}
&0.600
 \\

FMSC & NMI
&0.126
&\textbf{0.094}
&0.047
&0.211
&\underline{0.895}
&0.275
 \\

& BAL
&0.357
&\underline{0.305}
&\underline{0.432}
&0.412
&0.347
&0.371
 \\

& DPD $\downarrow$
&0.031
&0.110
&0.074
&0.017
&0.013
&0.049
 \\

\midrule
& ACC
&\textbf{0.465}
&\textbf{0.749}
&0.836
&0.864
&0.750
&\textbf{0.733}
 \\

AFMVC & NMI
&\underline{0.210}
&\textbf{0.094}
&0.047
&0.825
&0.840
&0.403
 \\

& BAL
&0.352
&0.302
&0.420
&0.440
&0.365
&\underline{0.376}
 \\

& DPD $\downarrow$
&0.028
&0.081
&0.054
&\textbf{0.009}
&\underline{0.010}
&0.036
 \\

\midrule
& ACC
&0.353
&0.589
&0.582
&0.878
&0.453
&0.571
 \\

FLFMVC & NMI
&0.162
&0.065
&0.046
&0.841
&0.594
&0.342
 \\

& BAL
&\textbf{0.384}
&\textbf{0.316}
&\textbf{0.433}
&0.445
&0.329
&\textbf{0.381}
 \\

& DPD $\downarrow$
&\textbf{0.008}
&\textbf{0.001}
&\textbf{0.000}
&\textbf{0.009}
&0.011
&\textbf{0.006}
 \\

\midrule
& ACC
&\underline{0.423}
&\underline{0.680}
&0.842
&\underline{0.968}
&0.709
&\underline{0.724}
 \\

Ours & NMI
&\textbf{0.217}
&\underline{0.071}
&0.049
&\underline{0.931}
&0.838
&\textbf{0.421}
 \\

& BAL
&\underline{0.379}
&0.286
&0.418
&\underline{0.446}
&\underline{0.367}
&\underline{0.379}
 \\

& DPD $\downarrow$
&\underline{0.017}
&0.080
&0.054
&\textbf{0.009}
&\textbf{0.009}
&0.034
 \\

\bottomrule
\end{tabular}
\end{table}
Multi-view fair clustering methods explicitly incorporate fairness considerations into multi-view representation learning or clustering optimization. FairMVC \cite{zheng2023fairness} promotes demographic parity by aligning the distribution of sensitive groups within individual clusters with their global distribution, while simultaneously enhancing multi-view representations through contrastive learning. FMSC \cite{li2024one} develops a one-stage spectral clustering framework equipped with a graph-based fairness regularizer, thereby integrating fair representation learning and clustering into a unified optimization procedure without requiring post-processing. AFMVC \cite{jiang2026adversarial} adopts an adversarial learning framework with a gradient reversal mechanism to reduce the dependence between learned clustering representations and sensitive attributes, thereby mitigating sensitive-information leakage and improving group fairness. FLFMVC \cite{zhao2026fairness} introduces a multi-level fairness mechanism that jointly refines base partitions, promotes balanced demographic distributions, and guides fair cluster assignments.

For all comparison methods, the hyperparameters are set according to the default configurations provided in their publicly available implementations. For our method, $\lambda_C$ and  $\lambda_F$ are both set to $0.1$. All experiments are conducted on a PC equipped with an Intel Core i7-10700F CPU at 2.90 GHz, 16 GB of RAM, and an NVIDIA GeForce RTX 1660 GPU with 6 GB of memory. For methods involving stochastic procedures, we independently repeat each experiment ten times and report the average results.
\subsection{Experimental Results}
Table~\ref{compare} reports the quantitative comparison among all competing methods. Based on these results, we highlight the following observations.
\begin{itemize}

\item Overall performance:
Overall, our method achieves a favorable trade-off between clustering utility and group fairness.
As reported in Table~\ref{compare}, Ours ranks first in terms of mean NMI and second in terms of both mean ACC and BAL.
These results demonstrate that the proposed method preserves strong clustering capability while maintaining a high level of fairness, yielding consistently balanced performance across different evaluation criteria.

\item Compared with single-view fair clustering methods:
Among the single-view fair clustering baselines, including BFKM, VFC, FFC, and FairDen, our method generally delivers superior overall performance, particularly in terms of clustering quality.
This advantage highlights the effectiveness of leveraging complementary information across multiple views while simultaneously preserving competitive group fairness.

\item Compared with conventional multi-view clustering methods:
Relative to conventional multi-view clustering approaches such as MCPL, CGL, and 3MC, our method achieves a more desirable trade-off between clustering quality and fairness.
Although some conventional multi-view methods attain strong clustering results on individual datasets, their advantages are less consistent when fairness is jointly considered.
In contrast, our method maintains competitive clustering accuracy while exhibiting more stable fairness performance across datasets.

\item Compared with multi-view fair clustering methods:
Among the fair multi-view clustering baselines, AFMVC achieves the highest mean ACC, whereas FLFMVC attains the strongest overall fairness performance.
In comparison, our method ranks first in mean NMI and second in both mean ACC and BAL, indicating a favorable balance between clustering performance and fairness.
Relative to AFMVC, our method improves group fairness at the cost of only a marginal decrease in ACC.
Compared with FLFMVC, our method retains substantially stronger clustering performance while still achieving competitive fairness.

\end{itemize}
\subsection{Ablation Study }
In this subsection, we assess the contribution of each loss component by removing them individually and analyzing the corresponding changes in clustering performance and fairness. The detailed results are presented in Table~\ref{ablation}, from which several observations can be drawn. First, removing either $\mathcal{L}_R$ or $\mathcal{L}_C$ generally leads to decreased ACC and NMI, confirming the importance of reconstruction learning and clustering supervision in preserving informative representations and maintaining clustering quality. Second, excluding either the asymmetric alignment loss $\mathcal{L}_A$ or the fairness regularization term $\mathcal{L}_F$ typically results in lower BAL and higher DPD, indicating that these two components enhance fairness from complementary perspectives. Specifically, $\mathcal{L}_A$ mitigates sensitive information at the representation level through asymmetric cross-view alignment, whereas $\mathcal{L}_F$ directly regularizes the consensus assignments to promote group-level fairness. Notably, removing $\mathcal{L}_F$ improves clustering performance on several datasets; however, such gains are accompanied by a clear deterioration in fairness, highlighting the inherent trade-off between clustering utility and fairness. Overall, the complete model achieves a more favorable balance by jointly incorporating all four objectives.
\begin{table}[pos=H]
\caption{Ablation study on different loss combinations. Each loss configuration is treated as an individual method. The best result for each metric is shown in bold.}
\label{ablation}
\centering
{
\begin{tabular}{c|cccccc}
\toprule
Loss & Metric  & Credit & Bank & Law  & Mfeat& COIL   \\
\midrule

& ACC &0.414 &0.659 &0.838 &0.923 &0.695 \\
w/o $\mathcal{L}_R$& NMI &0.198 &0.066 &0.047 &0.882 &0.835 \\

& BAL &0.368 &0.283 &0.406 &0.439&0.354 \\
& DPD $\downarrow$ &0.019 &0.091 &0.055 &0.009 &0.010 \\

\midrule

& ACC &0.417 &0.657 &0.824 &0.914 &0.675 \\
w/o $\mathcal{L}_C$
& NMI &0.193 &0.064 &0.045 &0.864 &0.819 \\

& BAL  &\textbf{0.381} &0.282 &\textbf{0.418} &\textbf{0.448} &0.363 \\
& DPD $\downarrow$ &\textbf{0.017} &0.100 &\textbf{0.053} &\textbf{0.008} &\textbf{0.009} \\

\midrule

& ACC &0.422 &0.663 &0.836 &\textbf{0.968} &0.756 \\
w/o $\mathcal{L}_A$
& NMI &0.217 &0.070 &0.047 &0.930 &\textbf{0.851} \\

& BAL &0.358 &0.283 &0.415 &0.439 &0.326 \\
& DPD $\downarrow$ &0.025 &0.094 &0.056 &0.010 &0.010 \\

\midrule

& ACC &\textbf{0.454} &0.676 &\textbf{0.845} &0.946 &\textbf{0.761} \\
w/o $\mathcal{L}_F$
& NMI &\textbf{0.229} &0.070 & \textbf{0.051}&0.909 &0.839 \\
& BAL &0.354 &0.282 &0.412 &0.438 &0.325 \\
& DPD $\downarrow$ & 0.027&0.097 &0.055 &0.010 &0.010 \\

\midrule

& ACC
&0.423
&\textbf{0.680}
&0.842
&\textbf{0.968}
&0.709
 \\

Full Model & NMI
&0.217
&\textbf{0.071}
&0.049
&\textbf{0.931}
&0.838
 \\

& BAL
&0.379
&\textbf{0.286}
&\textbf{0.418}
&0.446
&\textbf{0.367}
\\

& DPD $\downarrow$
&\textbf{0.017}
&\textbf{0.080}
&0.054
&0.009
&\textbf{0.009}
 \\

\bottomrule
\end{tabular}}
\end{table}
\subsection{Parameter Sensitivity Analysis}
\begin{figure}[pos=H]
\centering
\includegraphics[width=0.7\linewidth]{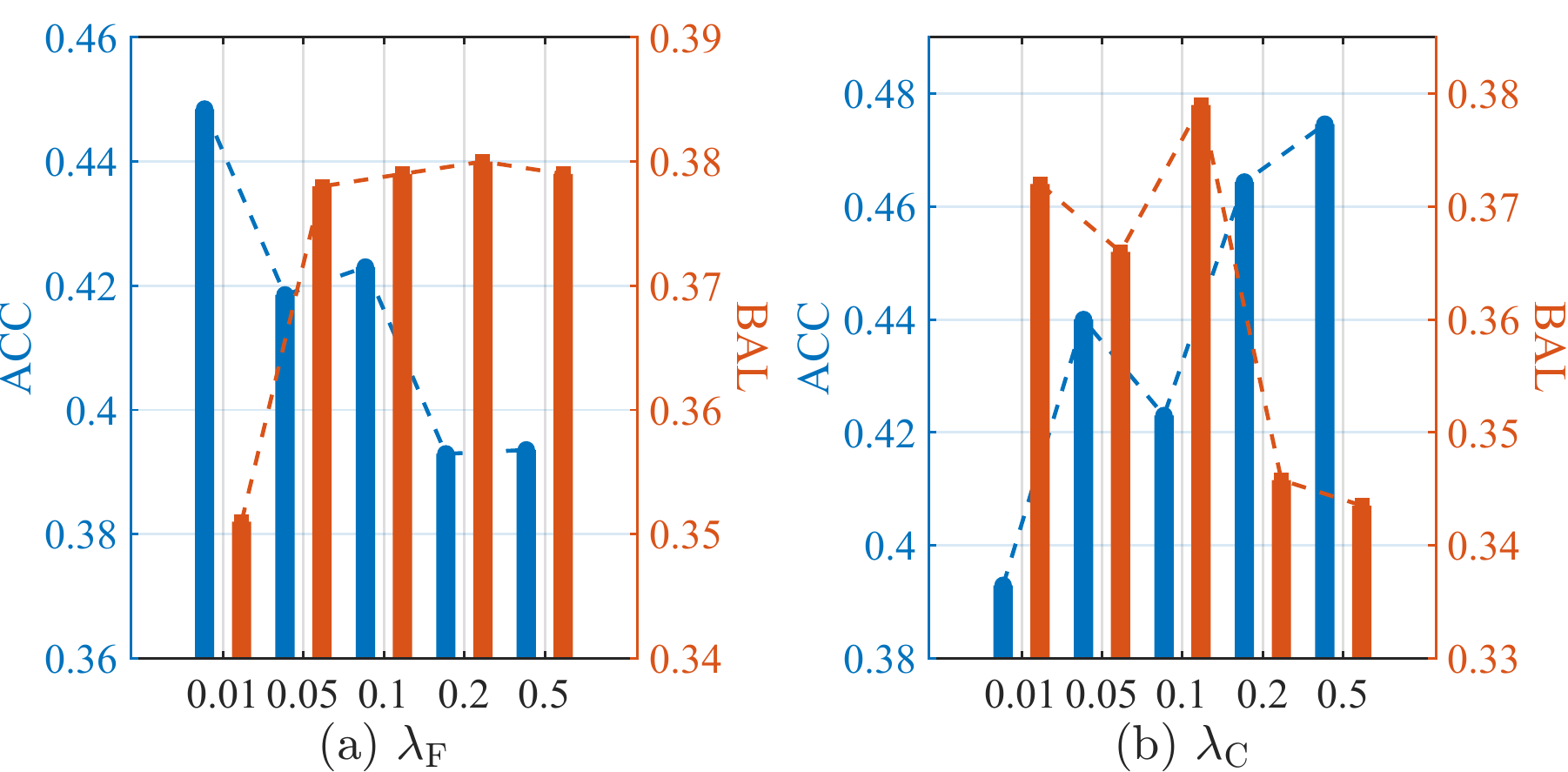} 
\caption{Parameter sensitivity analysis of $\lambda_F$ and $\lambda_C$ in terms of ACC and BAL.}
\label{fig:parameter}
\end{figure}

Our framework jointly optimizes reconstruction, clustering, asymmetric alignment, and fairness objectives, where $\lambda_C$ and $\lambda_F$ control the contributions of the clustering and fairness regularization terms, respectively.
To evaluate the sensitivity of these two hyperparameters, we vary one parameter at a time while fixing the other to its default value of $0.1$.
Specifically, both $\lambda_C$ and $\lambda_F$ are selected from $\{0.01, 0.05, 0.1, 0.2, 0.5\}$, and the corresponding ACC and BAL results are reported in Fig.~\ref{fig:parameter}.

As shown in Fig.~\ref{fig:parameter}, the two hyperparameters exhibit different effects on clustering performance and fairness.
When $\lambda_F$ increases from a relatively small value, BAL improves noticeably, while the fairness performance becomes relatively stable once $\lambda_F$ reaches a moderate level.
Further increasing $\lambda_F$ brings only marginal fairness gains while tending to reduce clustering accuracy, indicating that an excessively strong fairness constraint may compromise clustering utility.
For $\lambda_C$, a larger value generally favors clustering accuracy, whereas overemphasizing the clustering objective leads to a noticeable degradation in BAL.
These observations reveal a clear trade-off between clustering quality and fairness.
Overall, setting $\lambda_C=\lambda_F=0.1$ provides a favorable balance between the two objectives and is therefore adopted as the default configuration in our experiments.

\subsection{Robustness Analysis under Heterogeneous Sensitive Dependence}

To examine the effect of heterogeneous sensitive dependence across views, we construct controlled variants of COIL and Law with progressively increasing sensitive dependence in a selected target view.
For each dataset, we select an informative target view according to its single-view clustering performance and inject sensitive-attribute-related perturbations only into this view, while keeping all remaining views, cluster labels, and sensitive attributes unchanged.

After feature-wise standardization, let $\mathbf{x}_i$ denote the representation of the $i$-th sample in the target view.
We construct a sensitive perturbation $\boldsymbol{\delta}_i$ by centering the sensitive signal within each ground-truth cluster and constraining the perturbation direction to be orthogonal to the between-class discriminative subspace.
The perturbed representation is defined as
\[
\mathbf{x}_i^{\rho}
=
\mathcal{M}
\left(
\mathbf{x}_i
+
\sqrt{\rho}\,\boldsymbol{\delta}_i
\right),
\]
where $\rho$ controls the strength of the injected sensitive information and $\mathcal{M}(\cdot)$ preserves the overall feature scale.

The perturbation is designed to increase the sensitive dependence of the target view while minimizing interference with its original discriminative structure.
Based on the resulting sensitive-dependence levels, we use
$\rho\in\{0.001,0.005,0.01\}$ for COIL and
$\rho\in\{0.001,0.01,0.05\}$ for Law, corresponding to the Low, Medium, and High settings.

\begin{table}[pos=H]
\centering
\caption{Sensitive-dependence statistics under different levels of cross-view heterogeneity.}
\label{tab:heterogeneous_sensitive_dependence}
\small
\setlength{\tabcolsep}{5pt}
\renewcommand{\arraystretch}{1.15}
\begin{tabular}{c|cc|cc}
\toprule
\multirow{2}{*}{Setting}
& \multicolumn{2}{c|}{COIL}
& \multicolumn{2}{c}{Law} \\
\cmidrule(lr){2-3}
\cmidrule(lr){4-5}
& Target NHSIC & BiasGap
& Target NHSIC & BiasGap \\
\midrule
Original & 0.0087 & 0.0056 & 0.0090 & 0.0003 \\
Low      & 0.0205 & 0.0174 & 0.0150 & 0.0057 \\
Medium   & 0.0675 & 0.0644 & 0.0422 & 0.0328 \\
High     & 0.1257 & 0.1225 & 0.1378 & 0.1284 \\
\bottomrule
\end{tabular}
\end{table}

To verify the induced sensitive-information heterogeneity, we report two statistics for each setting in Table~\ref{tab:heterogeneous_sensitive_dependence}.
Target NHSIC denotes the dependence between the selected target view and the sensitive attribute, while BiasGap measures the difference between the maximum and minimum NHSIC values across all views.
Accordingly, Target NHSIC reflects the sensitive dependence of the perturbed target view, whereas BiasGap characterizes the resulting cross-view sensitive-information discrepancy.

As shown in Table~\ref{tab:heterogeneous_sensitive_dependence}, both Target NHSIC and BiasGap increase monotonically from the Original to the High setting on COIL and Law.
The increasing Target NHSIC confirms that the sensitive dependence of the selected view is progressively strengthened, while the corresponding increase in BiasGap demonstrates that the discrepancy in sensitive dependence across views becomes increasingly pronounced.
These results validate the constructed variants as controlled settings with progressively stronger cross-view sensitive-information heterogeneity.

\begin{figure*}[pos=h]
\centering
\includegraphics[width=1\linewidth]{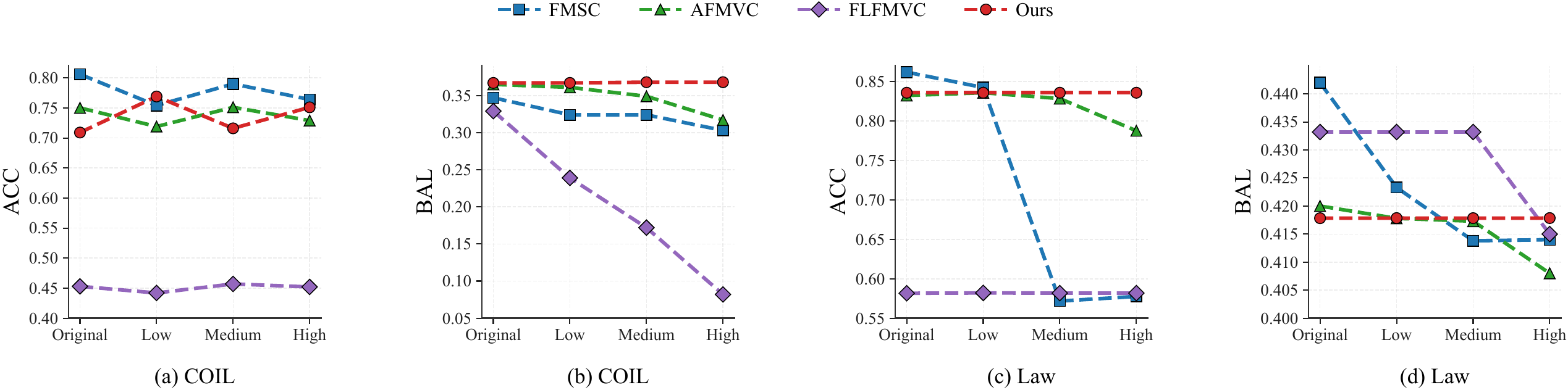}
\caption{Comparison of multi-view fair clustering methods under increasing cross-view sensitive-information heterogeneity.}
\label{case}
\end{figure*}

Based on these controlled variants, we further compare the clustering performance and fairness of different methods as cross-view sensitive-information heterogeneity increases.
The results are shown in Fig.~\ref{case}.
On COIL, our method maintains an almost unchanged BAL score as the sensitive dependence of the target view increases, whereas the competing methods exhibit more noticeable fairness degradation, especially under the High setting.
Meanwhile, its ACC remains competitive across all perturbation levels.
A similar but more pronounced pattern is observed on Law.
The ACC and BAL of our method remain nearly unchanged from the Original to the High setting, whereas the competing methods become increasingly affected by the injected sensitive information.
In particular, AFMVC exhibits a clear decline in both clustering utility and fairness under the High-heterogeneity setting, while FMSC shows a substantial decrease in ACC as the perturbation strength increases.

Overall, these results indicate that explicitly modeling view-specific sensitive dependence and regulating cross-view interactions improves robustness to heterogeneous sensitive information.
The proposed method therefore maintains a more stable balance between clustering quality and fairness as the sensitive-information discrepancy across views increases.

\section{Conclusion}
\label{5}
In this paper, we propose a novel multi-view  fair clustering framework guided by cross-view sensitive information discrepancy.
The proposed method explicitly estimates the sensitive dependence of different views and employs a bias-ranked asymmetric alignment mechanism, encouraging views with higher sensitive dependence to learn from those with lower sensitive dependence while adaptively regulating the alignment process according to cross-view discrepancy.
In addition, fairness regularization is imposed on the consensus soft assignments to further promote group fairness at the clustering-assignment level.
By jointly considering cross-view sensitive-information heterogeneity, representation alignment, and assignment-level fairness, the proposed framework effectively balances clustering accuracy and group fairness.
Experimental results on five benchmark datasets demonstrate competitive clustering performance together with strong group fairness.
Moreover, the robustness analysis under heterogeneous sensitive dependence further confirms that our method can better cope with cross-view sensitive-information heterogeneity than existing multi-view fair clustering approaches, particularly in maintaining stable fairness as the sensitive dependence of the target view increases.

In future work, we plan to investigate more scalable strategies for estimating and regulating cross-view sensitive information, and further extend the framework to more complex scenarios, such as continuous sensitive attributes, incomplete multi-view data, and streaming multi-view settings.

\section*{Acknowledgments}
This work has been supported by the  National Natural Science Foundation of China under Grant Nos. 62476038 and 62472064. 
\bibliographystyle{cas-model2-names}

\bibliography{cas-refs}



\end{document}